\documentclass[runningheads]{llncs}
\usepackage[T1]{fontenc}
\usepackage{graphicx,verbatim}
\usepackage{amsmath}
\usepackage{amssymb}
\usepackage{booktabs}
\usepackage{multirow}
\usepackage{array}
\begin{document}
\title{CEVAR: Centerline Embedding Extraction for Endovascular Aneurysm Repair}

\titlerunning{CEVAR}
%
\author{Roman Naeem\inst{1} \and
Timo Niiniskorpi\inst{2} \and
Charlotte Sandström\inst{2} \and
Naman Desai\inst{1} \and
Anders Jeppsson\inst{3} \and
Ida Häggström\inst{1} \and
Fredrik Kahl\inst{1} \and
Håkan Roos\inst{2} \and
Jennifer~Alvén\inst{1}}
\authorrunning{F. Author et al.}
%
\institute{Chalmers University of Technology, 412 96 Gothenburg, Sweden \and
The University of Gothenburg, 405 30 Gothenburg, Sweden  \and
Sahlgrenska University Hospital, 413 45 Gothenburg, Sweden}
  
\maketitle              
\begin{abstract}

Long-term mortality rates after endovascular aneurysm repair (EVAR) remain elevated due to post-EVAR rupture caused by loss of seal in stent graft sealing zones. Structured CT review using centerline measurements improves detection, but current workflows require manual centerline editing and expert operators. We propose a transformer framework for automated, protocol-driven sealing zone assessment that combines 3D centerline tracking with embedding-based geometric prediction. Two state-of-the-art image-to-graph models are evaluated for aorto-iliac centerline extraction from follow-up CT and for measurement of stent position, vessel diameters, and seal lengths according to EVAR4C protocol. Across the full test set and a challenging no-contrast subset, the proposed fully automatic method outperforms the commercial semi-automatic workflow. Code and pretrained models are available at \url{https://github.com/RomStriker/CEVAR}. 


\end{abstract}

\section{Introduction}

Abdominal aortic aneurysm is a dilation of the abdominal aorta exceeding 50\% of normal diameter ~\cite{johnston1991suggested}. Enlargement increases rupture risk and treatment is recommended beyond a threshold size~\cite{esvs_guidelines_2024}. Endovascular Aneurysm Repair (EVAR) is now the predominant treatment due to favorable short term outcomes, yet long term survival remains limited due to late post-EVAR rupture~\cite{powell2017meta}.

Post-EVAR rupture is strongly associated with progressive loss of seal in proximal or distal stent graft sealing zones~\cite{diehm2008aortic,gonccalves2017iliac}. Structured follow-up protocols detect loss of seal in up to 40\% of patients within five years and predict failure more reliably than aneurysm sac expansion or endoleaks~\cite{SANDSTROM2025238}.

The EVAR4C protocol~\cite{SANDSTROM2025238} evaluates sealing zones using vessel centerlines as reference for length and orthogonal diameter measurements. It more than doubles the detection of seal loss compared to conventional review, but requires manual centerline editing by highly trained operators~\cite{andersson2022structured,SANDSTROM2025238}.

Accurate centerline generation is a major bottleneck. Existing tools are designed for contrast-enhanced preoperative CT and require extensive manual adjustment~\cite{adam2021pre,corriere2014influence} in post-EVAR scans affected by metal artifacts, altered anatomy, and low contrast.

Full automation of the EVAR4C protocol would enable standardized, objective sealing-zone evaluation in routine clinical follow-up, reducing operator dependency and manual workload. Achieving this requires reliable centerline extraction and protocol-specific measurements. We evaluate Transformer-based image-to-graph methods for centerline prediction in post-EVAR CT and investigate whether learned embeddings enable estimation of clinically relevant metrics.

Our key contributions are:
\begin{itemize}
\item A systematic assessment of Transformer-based 3D centerline extraction in post-EVAR CT, addressing challenges of metal artifacts and suboptimal contrast.
\item A curated clinical dataset featuring manually refined centerlines and expert annotations for stent edges and sealing-zone measurements.
\item An embedding-based framework that leverages learned point-wise representations for multi-task prediction of stent classification, vessel radii, and seal-zone length.
\item An automated, protocol-driven pipeline for centerline and geometric quantification, marking a first step toward automated sealing-zone-focused post-EVAR follow-up.
\end{itemize}

\begin{figure}[!t]
\centering
\includegraphics[width=1.0\textwidth]{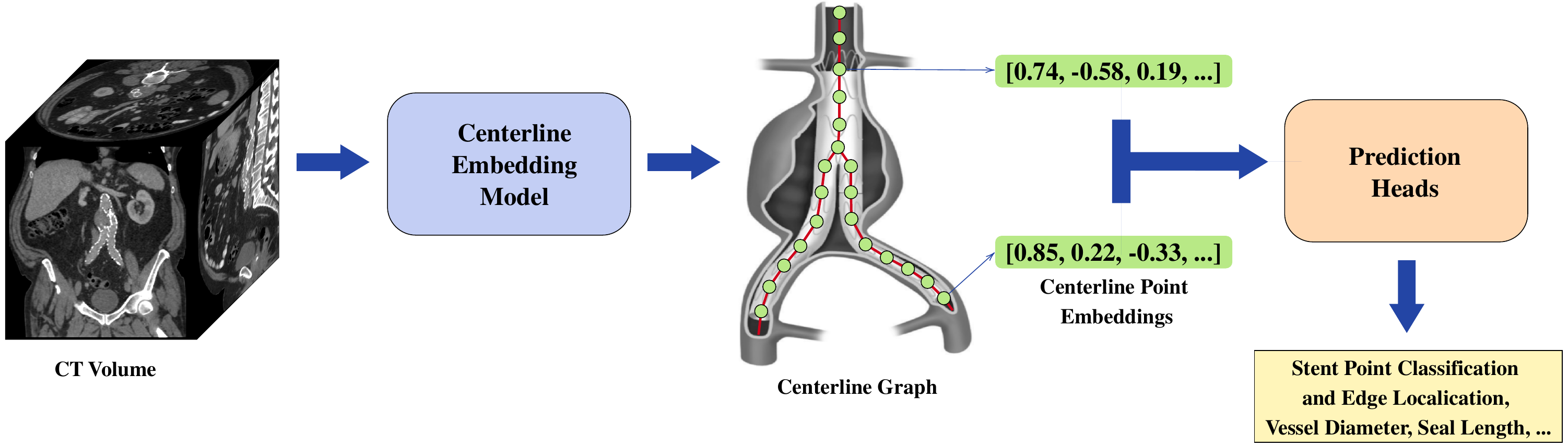}
\caption{
Proposed pipeline for automated post-EVAR analysis from CT. A CT volume is processed by a Transformer-based 3D centerline model that outputs the vessel centerline as point-wise embeddings. These embeddings are fed into a feature-prediction network to estimate clinically relevant measurements, including stent-region labels, vessel radii, and sealing-zone lengths, enabling automated EVAR4C protocol assessment.
}
\label{fig:arch}
\end{figure}

\subsubsection{Related Work.}

Deep learning methods for EVAR imaging address screening~\cite{spinella2023artificial}, endoleak detection~\cite{yang2024detection}, morphological change~\cite{lareyre2025imaging}, migration~\cite{asenbaum2019stent}, and risk stratification~\cite{long2024applying}. These approaches detect late manifestations of failure and do not quantify sealing zone integrity. Prior work~\cite{sabrowsky2021automatic} on centerline extraction produces segmentations rather than protocol measurements. To our knowledge, no previous study has used learned centerline embeddings for automated sealing zone quantification in post-EVAR CT.

\section{Method}
Postoperative CT examinations after EVAR were evaluated according to the EVAR4C protocol for CT review. The protocol specifies centerline-based measurements along each stent-graft limb, including sealing-zone lengths and orthogonal diameters at predefined locations. Vessel and stent-graft diameters were measured at (i) the stent-graft edges, (ii) 10 and 15 mm from each stent-graft edge, and (iii) the sealing-zone margin closest to the aneurysm. Sealing lengths were measured along the centerline and annotated for each graft limb. 

\subsection{Datasets}
\subsubsection{Private Dataset.} The dataset was collected retrospectively from a clinical cohort of patients treated with standard bifurcated EVAR at two hospitals during 2010-2012. Ethical approval was granted (508-14). For each patient, the preoperative CT, the postoperative CT at one month, the postoperative CT at one year, and the last available follow-up CT were reviewed. For each CT examination, centerlines were generated and subsequently manually adjusted by the reviewing physician, using the commercially avaliable software Terarecon Intuition~\cite{terarecon}.  Markers indicating the stent-graft edges were placed manually. All measurements and annotations were performed according to the EVAR4C protocol. 

To train the centerline extraction models, we utilized 374 scans from 142 subjects, partitioned by subject into training, validation, and test sets. The training set contains 267 scans (102 subjects; 207 contrast/60 non-contrast), the validation set 32 scans (12 subjects; 30 contrast/2 non-contrast), and the test set 75 scans (28 subjects; 59 contrast/16 non-contrast).

\subsubsection{Aorta24 Centerline Dataset.} We derived a centerline dataset from the public Aorta24 challenge~\cite{imran2024multiclass}, which consists of 100 CT scans with corresponding multiclass aorta segmentation masks. Centerlines were automatically extracted from these masks using the Vascular Modeling Toolkit (VMTK)~\cite{Izzo2018,slicer3d}. This processed dataset served as the foundation for pretraining the centerline models.

\subsection{Centerline Extraction Methods}
We compare the commercial postprocessing workflow with state-of-the-art deep learning-based image-to-graph centerline extraction methods.

\subsubsection{TeraRecon Intuition.}
TeraRecon Intuition~\cite{terarecon} is a standard commercial platform for radiological post-processing. Its dedicated EVAR workflow uses an intensity-guided algorithm that assumes sufficient contrast enhancement for vessel tracking. In this study, we evaluate the software using three distinct initialization modes to establish a comprehensive baseline.

The automatic (SemAuto) mode generates multiple candidate paths, requiring the user to manually select the correct left and right aortic centerlines. In single-seed (2Seeds) mode, a seed is placed at the proximal aorta to generate a centerline for one branch, and a second seed is placed at the distal end of the other iliac branch to connect it to the first centerline. Finally, the two-click (4Seeds) mode requires the user to mark the vessel above and below the region of interest for both aorto-iliac branches. In all configurations, the final centerline is computed by linking these user-defined markers based on local voxel intensities.

\subsubsection{Trexplorer Super.}
Trexplorer Super~\cite{naeem2025trexplorer} is an image-to-graph framework using a Transformer architecture \cite{naeem2024trexplorer,vaswani2017attention} to extract vascular centerline graphs from CT volumes via sequential tracking. From a localized seed point, the model expands the centerline graph breadth-first, ensuring a topologically consistent tree without disconnected components or cycles. Its core is a recursive tracking module that autoregressively predicts the next point embedding from the current point, capturing both local image features and long-range spatial dependencies along the vessel lumen.

Tracking is guided by a multi-class classification head labeling embeddings as intermediate, bifurcation, end, or background. Tracking proceeds along intermediate points and stops at end or background points. Bifurcations spawn multiple embedding copies to initialize potential new branches; some continue tracking while others are discarded based on subsequent classifications. This mechanism enables the model to dynamically follow complex, multi-scale vascular hierarchies while preserving global connectivity.

\subsubsection{RefTr.}
RefTr~\cite{naeem2026reftr} is a parameter-efficient 3D image-to-graph model representing centerline trees as confluent trajectories that share a root and follow the same path until bifurcation. Using a Transformer-based Producer–Refiner architecture, the Producer generates candidate trajectories from the input patch root at its center, and a shared Refiner iteratively aligns them with the target branches. This approach refines entire trajectories while enforcing topologically valid tree structures, improving precision and making the model lighter and faster than Trexplorer Super.

To boost accuracy, RefTr predicts multiple candidates per target, creating an ensembling effect resolved via a novel Tree Non-Maximum Suppression (TNMS) algorithm that merges redundant branches while preserving global topology. It also extends Trexplorer Super’s evaluation framework, supporting point-, branch-, and graph-level metrics for centerline assessment, which we adopt in this work.

\subsubsection{Prediction Heads.}
Due to RefTr’s superior efficiency and performance, we extend it to predict EVAR4C protocol measurements. Beyond the standard centerline embedding heads, we add:

\begin{enumerate}
\item \textbf{Stent point classification head:} A binary head that predicts whether a point is inside or outside the stent, used to localize proximal and distal stent edges for diameter and seal length measurements.
\item \textbf{Stent edge regression head:} For stent edge points, a learned token is introduced, which is processed to estimate four values: edge diameter, diameters at 10mm and 15mm inward, and seal length.
\end{enumerate}

\subsubsection{Automatic Seed Localization.} To enable a fully automated post-EVAR analysis pipeline, we replace manual initialization with a deep-learning-based seed localization module. A pretrained SwinUNETR \cite{hatamizadeh2021swin} was fine-tuned for 80,000 iterations to regress a heatmap centered on the stent’s proximal edge. The seed point is then extracted by thresholding the heatmap and selecting its peak coordinates, which serve as the starting point for both Trexplorer Super and RefTr.

\subsubsection{Training Strategy.}
Trexplorer Super and RefTr were pretrained on the Aorta24 centerline dataset for 2 million iterations. These pretrained weights initialized all subsequent fine-tuning experiments on the private dataset, with each model fine-tuned for 200,000 iterations. Training used a cosine annealing learning rate scheduler with a single 20,000-iteration warm-up at the start and a base learning rate of $7 \times 10^{-4}$, on a single NVIDIA RTX A6000 GPU (48GB VRAM) with a batch size of 8. Pretraining required 190 GPU hours for Trexplorer Super and 124 GPU hours for RefTr, while fine-tuning took 21 GPU hours for Trexplorer Super and 14 GPU hours for RefTr.

\section{Experiments and Results}
\subsection{Evaluation Metrics}
For centerline evaluation, we adopt the hierarchical metrics proposed in RefTr \cite{naeem2026reftr}. Node-level performance is measured using radius-aware Average Precision ($rAP$), Recall ($rAR$), and F1-score ($rF1$). A predicted node is considered a true positive (TP) if it lies within a radius $\tau^{rad}$ of an unmatched target node. These metrics are averaged over $\tau^{rad} \in [3.0, 15.0]$ mm with a 1.5 mm step. Branch-level metrics ($rBAP$, $rBAR$, $rBF1$) evaluate structural connectivity, where a branch is a TP if it covers at least a fraction $\tau^{match}$ of nodes from an unmatched target branch. These are averaged over $\tau^{match} \in [0.4, 0.8]$ with a 0.05 step, using $\tau^{rad} = 6.0$ mm. Predictions outside the stent region are discarded, as the ground truth centerline is not well-defined in these areas. Graph-level topology is assessed using the Mean Absolute Error (MAE) of Betti-0 (connected components) and Betti-1 (cycles) numbers. We also report MAE for vessel diameter and seal length.

For stent node classification, we report mean Average Precision (mAP), mean Recall (mRec), and mean maximum F1-score (mF1), averaged across all $\tau^{rad}$ thresholds. All results reflect the mean and standard deviation across three independent runs.

\begin{table}[!t]
\centering
\caption{Comparison of TeraRecon (TR), Trexplorer Super (Trex), Trexplorer Super with Tree Non-maximum Suppression (TrexTNMS), and RefTr under different seeding strategies (SemAuto, 2Seeds, 4Seeds, 1Seed, Auto), evaluated using point-level and branch-level metrics for the \textbf{full test set}. Best scores are underlined.} 
\label{tab:res-table-1}
{\fontsize{6}{8}\selectfont
\begin{tabular}{c|c|c|c|c|c|c}
\toprule
\multirow{2}{*}{Model} 
& \multicolumn{3}{c|}{Point-level${@\,\tau^{\text{rad}}=[3.0{:}1.5{:}15.0]mm}$}
& \multicolumn{3}{c}{Branch-level${@\,\tau^{\text{match}}=[0.4{:}0.05{:}0.8]}$} \\

& rAP(\%)$\uparrow$ & rAR(\%)$\uparrow$ & rF1(\%)$\uparrow$
& rBAP(\%)$\uparrow$ & rBAR(\%)$\uparrow$ & rBF1(\%)$\uparrow$ \\
\midrule

\multicolumn{7}{c}{Commercial Semi-automatic Methods} \\
\midrule
TR:SemAuto & 2.01 & 2.27 & 2.11 & 2.67 & 2.27 & 2.37 \\
TR:2Seeds  & 53.48 & 58.56 & 55.19 & 71.83 & 60.59 & 63.90 \\
TR:4Seeds  & 60.29 & 66.06 & 62.64 & 76.77 & 66.27 & 68.91 \\

\midrule
\multicolumn{7}{c}{Centerline Semi-automatic Methods (1Seed)} \\
\midrule
Trex   & 68.61 ± 1.07 & 72.38 ± 1.07 & 69.12 ± 0.23 & 89.76 ± 0.95 & 80.64 ± 2.52 & 83.27 ± 1.18 \\
TrexTNMS  & 71.24 ± 0.95 & 72.99 ± 1.77 & 70.78 ± 0.30 & 94.29 ± 1.31 & 79.95 ± 2.00 & 84.57 ± 0.39 \\
RefTr  & 70.49 ± 0.22 & \textbf{78.53 ± 0.59} & 73.27 ± 0.23 & \textbf{94.52 ± 0.91} & \textbf{85.50 ± 0.94} & \textbf{88.47 ± 0.38} \\

\midrule
\multicolumn{7}{c}{Centerline Fully Automatic Methods (Auto)} \\
\midrule
Trex   & 74.83 ± 1.67 & 71.50 ± 0.91 & 71.33 ± 0.68 & 89.11 ± 2.44 & 59.36 ± 0.66 & 68.58 ± 0.27 \\
TrexTNMS  & \textbf{78.47 ± 1.39} & 72.49 ± 1.99 & 73.61 ± 0.35 & 94.49 ± 2.25 & 60.15 ± 1.20 & 70.26 ± 0.74 \\
RefTr  & 76.12 ± 0.29 & 75.76 ± 0.92 & \textbf{74.59 ± 0.65} & 93.75 ± 0.80 & 66.19 ± 1.10 & 75.23 ± 0.66 \\

\bottomrule
\end{tabular}}
\end{table}

\begin{table}[!t]
\centering
\caption{Comparison of TR, Trex, TrexTNMS, and RefTr across different seeding strategies (SemAuto, 2Seeds, 4Seeds, 1Seed, Auto), evaluated with point-level and branch-level metrics on the \textbf{non-contrast test set}. Best scores are underlined.}
\label{tab:res-table-2}
{\fontsize{6}{8}\selectfont
\begin{tabular}{c|c|c|c|c|c|c}
\toprule
\multirow{2}{*}{Model} 
& \multicolumn{3}{c|}{Point-level${@\,\tau^{\text{rad}}=[3.0{:}1.5{:}15.0]mm}$}
& \multicolumn{3}{c}{Branch-level${@\,\tau^{\text{match}}=[0.4{:}0.05{:}0.8]}$} \\

& rAP(\%)$\uparrow$ & rAR(\%)$\uparrow$ & rF1(\%)$\uparrow$
& rBAP(\%)$\uparrow$ & rBAR(\%)$\uparrow$ & rBF1(\%)$\uparrow$ \\
\midrule

\multicolumn{7}{c}{Commercial Semi-automatic Methods} \\
\midrule
TR:SemAuto & 0.00 & 0.00 & 0.00 & 0.00 & 0.00 & 0.00 \\
TR:2Seeds  & 4.90 & 4.07 & 4.08 & 7.41 & 4.17 & 4.91 \\
TR:4Seeds  & 15.20 & 13.56 & 14.20 & 25.69 & 13.19 & 13.21 \\

\midrule
\multicolumn{7}{c}{Centerline Semi-automatic Methods (1Seed)} \\
\midrule
Trex   & 66.94 ± 1.89 & 60.29 ± 4.50 & 62.31 ± 2.73 & 90.28 ± 3.64 & 66.44 ± 7.74 & 73.24 ± 7.23 \\
TrexTNMS  & 68.47 ± 1.37 & 60.83 ± 4.84 & 63.35 ± 3.04 & \textbf{91.20 ± 4.07} & 66.67 ± 7.50 & \textbf{73.77 ± 7.28} \\
RefTr  & 58.42 ± 2.24 & 59.87 ± 1.22 & 58.10 ± 0.60 & 83.02 ± 1.36 & \textbf{69.21 ± 2.73} & 73.00 ± 1.60 \\

\midrule
\multicolumn{7}{c}{Centerline Fully Automatic Methods (Auto)} \\
\midrule
Trex   & 76.33 ± 0.60 & 61.00 ± 1.76 & 66.61 ± 0.74 & 88.66 ± 2.82 & 46.53 ± 0.93 & 58.18 ± 1.02 \\
TrexTNMS  & \textbf{77.84 ± 0.49} & \textbf{61.96 ± 2.32} & \textbf{67.84 ± 1.18} & 90.82 ± 1.67 & 47.69 ± 1.22 & 59.92 ± 0.84 \\
RefTr  & 68.69 ± 2.65 & 60.99 ± 1.77 & 63.38 ± 0.58 & 85.26 ± 5.08 & 53.01 ± 1.01 & 62.59 ± 1.11 \\

\bottomrule
\end{tabular}}
\end{table}

\subsection{Results}
Table~\ref{tab:res-table-1} summarizes point- and branch-level performance on the full test set. Commercial semi-automatic extraction using TeraRecon (TR) required multiple seeds to achieve reasonable accuracy but remained below the performance of learning-based methods. Among single-seed approaches, RefTr achieved the best overall performance, with the highest point-level recall and F1, as well as branch F1. Applying Tree Non-max Suppression to Trexplorer Super (TrexTNMS) improved branch precision, but it still lagged behind RefTr in recall.

In the fully automatic setting, all learning-based methods outperformed the commercial baseline. TrexTNMS achieved the highest precision, while RefTr provided the best precision–recall balance, yielding the highest branch F1 and recall. This indicates superior recovery of complete vascular trees without manual initialization.

Results on the non-contrast subset (Table~\ref{tab:res-table-2}) reveal severe degradation for the commercial method, with near-zero scores even with additional seeds. Learning-based methods remained robust, with TrexTNMS achieving the highest fully automatic point-level F1, and RefTr showing competitive branch recall, suggesting improved branch recovery under low-contrast conditions.

On the full test set, commercial extraction showed decreasing connectivity errors with more seeds (Betti-0 MAE: 0.960 $\rightarrow$ 0.347 $\rightarrow$ 0.027), whereas all learning-based methods achieved perfect topology (Betti-0 = Betti-1 = 0). On non-contrast scans, the commercial method degraded noticeably (Betti-0 MAE: 0.938 $\rightarrow$ 0.812 $\rightarrow$ 0.063), while learning-based approaches again produced topologically correct trees in all modes.

\begin{figure*}[!t]
\centering
\includegraphics[width=0.99\textwidth]{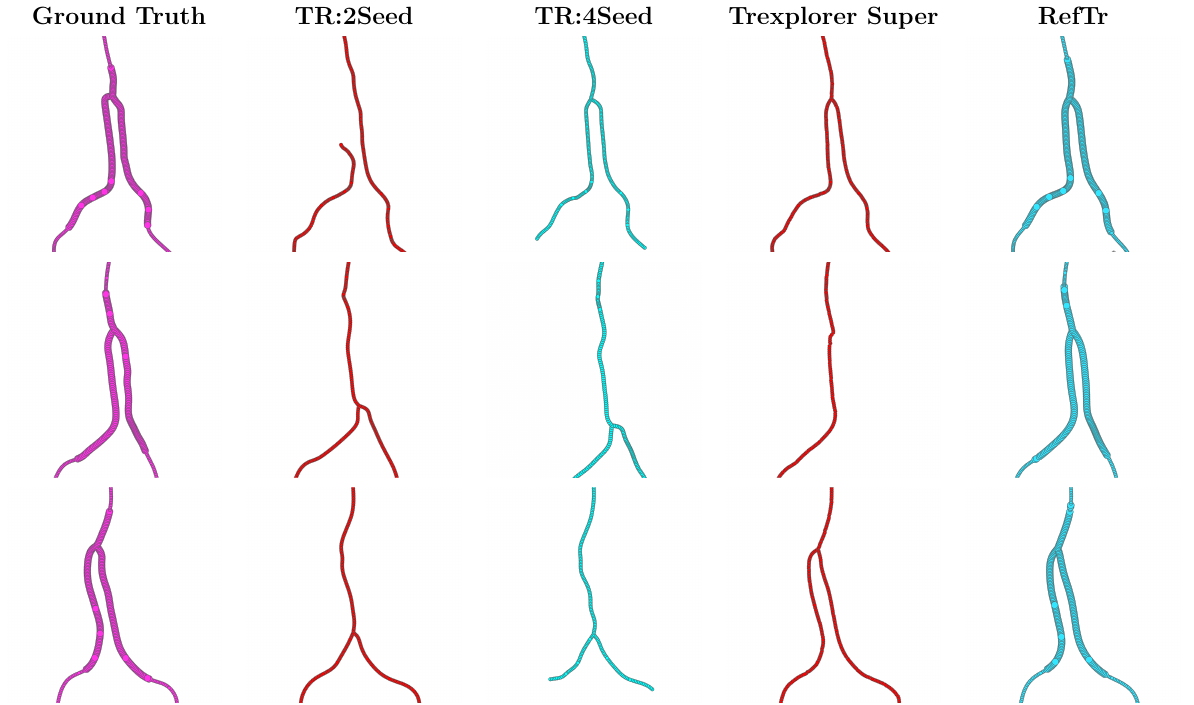}
\caption{Visual comparison of ground truth and predicted centerlines. Larger points indicate stent nodes for the ground truth and RefTr, which uses a stent classification head. TR:AutoSem is omitted due to poor performance.}
\label{fig:comp}
\end{figure*}

Vessel diameter measurements across proximal and distal stent edges ranged from approximately 7.6 to 56.5 mm, reflecting substantial anatomical variability. Seal lengths exhibited an even wider distribution, from 0 to 77.6 mm across sealing-zones. Table~\ref{tab:res-table-3} compares diameter and seal length measurement accuracy. Fine-tuning the pretrained radius head achieved lower diameter errors than the additional-token approach, whereas the additional token enabled seal-length prediction at with larger error. Stent-region classification showed strong performance across models (mAP = 0.860 ± 0.003, mRec = 0.763 ± 0.008, mF1 = 0.826 ± 0.005). Qualitative examples of these results are shown in Figure~\ref{fig:comp}.

\begin{table}[!t]
\centering
\caption{Comparison of two approaches for diameter prediction: (1) fine-tuning the pretrained radius prediction head, and (2) introducing an additional token with a new prediction head. Diameters are evaluated at the three stent edges, as well as at 10 mm and 15 mm below the proximal edge and above the left and right distal edges. Stent seal length cannot be predicted using the pretrained head and is therefore reported only for the additional-token method.}
\label{tab:res-table-3}
{\fontsize{6}{8}\selectfont
\begin{tabular}{c|c|c|c|c|c|c}
\toprule
\multirow{2}{*}{Model} 
& \multicolumn{3}{c|}{Proximal Edge Diameter (MAE)$\downarrow$} 
& \multicolumn{3}{c}{Right Edge Diameter (MAE)$\downarrow$} \\

& Edge & @10mm & @15mm
& Edge & @10mm & @15mm \\
\midrule
Radius Finetune  & \textbf{1.27 ± 0.10} & \textbf{1.80 ± 0.06} & \textbf{2.07 ± 0.13} & \textbf{2.56 ± 0.11} & \textbf{1.81 ± 0.10} & \textbf{1.94 ± 0.05} \\
Additional Token  & 1.59 ± 0.15 & 2.43 ± 0.09 & 2.72 ± 0.05 & 3.49 ± 0.40 & 3.25 ± 0.59 & 3.22 ± 0.65 \\
\midrule
\multirow{2}{*}{Model} 
& \multicolumn{3}{c|}{Left Edge Diameter (MAE)$\downarrow$} 
& \multicolumn{3}{c}{Stent Seal Length (MAE)$\downarrow$} \\
& Edge & @10mm & @15mm
& Proximal & Right & Left \\
\midrule
Radius Finetune  & \textbf{2.14 ± 0.08} & \textbf{1.80 ± 0.01} & \textbf{2.20 ± 0.13} & - & - & - \\
Additional Token  & 3.05 ± 0.19 & 2.90 ± 0.11 & 3.05 ± 0.15 & \textbf{9.10 ± 0.38} & \textbf{13.12 ± 0.54} & \textbf{11.90 ± 0.23} \\

\bottomrule
\end{tabular}}
\end{table}

\section{Discussion and Conclusion}
In this work, we propose a Transformer-based framework for automated and protocol-driven sealing zone assessment after EVAR, combining 3D centerline tracking with embedding-based geometric prediction.  Compared to commercially available semi-automatic software, our approach demonstrates substantially improved geometric accuracy and topological consistency, including in challenging non-contrast CT examinations.

Accurate centerline extraction is a prerequisite for reliable sealing zone evaluation within the EVAR4C protocol. The results demonstrate that learning-based centerline tracking can reduce dependence on manual adjustment while preserving the geometric consistency required for downstream measurements. In particular, the learned point-wise centerline embeddings enable reliable identification of stent graft edges and accurate diameter measurements at predefined protocol locations. Seal length prediction remains challenging, likely reflecting the difficulty of modeling vessel--stent interactions, especially where separation between vessel wall and stent graft is subtle, and improved modeling likely requires larger training cohorts and richer geometric supervision. Ongoing work therefore focuses on expanding the cohort and incorporating more comprehensive pixel-wise annotations of sealing zone geometry.

Importantly, the proposed framework enables the use of existing large-scale longitudinal follow-up data consisting of routine CT examinations and associated tabular measurements, which opens the possibility of developing and validating fully automated EVAR follow-up pipelines at substantially larger scale. Overall, the results indicate that Transformer-based centerline tracking combined with embedding-based geometric prediction can provide a robust foundation for automated, protocol-driven post-EVAR evaluation.

\begin{credits}
\subsubsection{\ackname} The compute resources were provided by the National Academic Infrastructure for Supercomputing in Sweden (NAISS), partially funded by the Swedish Research Council through grant agreement no. 2022-06725.

\subsubsection{\discintname} The authors have no competing interests to declare that are relevant to the content of this article.
\end{credits}

%
%
%
\bibliographystyle{splncs04}
\bibliography{Paper}

\end{document}